\documentclass[11pt]{article}

\usepackage[final]{acl}
\usepackage{times}
\usepackage{latexsym}
\usepackage[T1]{fontenc}
\usepackage[utf8]{inputenc}
\usepackage{microtype}
\usepackage{inconsolata}
\usepackage{graphicx}
\graphicspath{{figures/}}
\usepackage{amsmath}
\usepackage{amssymb}
\usepackage{booktabs}
\usepackage{multirow}
\usepackage{hyperref}

\title{The Temporal Tug-of-War: Visualizing and Detecting RAG Conflicts in Diffusion Models via Trajectory Variance}

\author{
\textbf{Sravan Karthick T}$^{1}$ \quad
\textbf{Pranav Darshan}$^{1}$ \quad
\textbf{Pranav A}$^{1}$ \\
\textbf{Dr. Minal Moharir}$^{1}$ \quad
\textbf{Dr. Ivan P. Yamshchikov}$^{2}$ \\[4pt]
$^{1}$Department of Computer Science and Engineering, R.V. College of Engineering, India \\
$^{2}$CAIRO, Technical University of Applied Sciences Würzburg-Schweinfurt, Germany \\[4pt]
\texttt{sravankt.cs20@rvce.edu.in, pranavdarshan.cs22@rvce.edu.in} \\
\texttt{pranava.cs21@rvce.edu.in, minalmoharir@rvce.edu.in} \\
\texttt{ivan.yamshchikov@thws.de}
}

\begin{document}

\maketitle

\begin{abstract}
Retrieval-Augmented Generation (RAG) introduces a specific failure mode in discrete diffusion language models: when retrieved context contradicts parametric knowledge, the iterative denoising process becomes a visible battleground between competing knowledge sources. We identify temporal semantic divergence as an observable for detecting these conflicts and introduce the Trajectory Variance Score (TVS), a simple and interpretable measure of this divergence. TVS computes the mean pairwise cosine distance of answer embeddings across independent stochastic denoising trajectories, capturing the temporal tug of war between parametric and contextual attractors. Requiring as few as two parallel inference runs, TVS is computationally lightweight. Across four diverse datasets (Synthetic, SciQ, PopQA, and CounterFact), a simple Logistic Regression classifier using TVS achieves $70.10\%$ accuracy and $0.7647$ AUROC on LLaDA. On Dream 7B, increasing the number of trajectories from two to five improves accuracy from $63.91\%$ to $69.62\%$. More complex sequential models provide only marginal improvements over the linear classifier. Evaluation across LLaDA and Dream 7B demonstrates that conflict-induced trajectory dynamics and their key properties transfer across distinct diffusion architectures.
\end{abstract}

\section{Introduction}
Using Retrieval Augmented Generation (RAG) \citep{lewis2020retrieval} is now the standard way to reduce hallucinations in large language models. However, this design introduces a secondary failure mode known as knowledge conflict. When an injected context directly contradicts the facts the model learned during training (such as outdated information or adversarially generated fakes), the model must implicitly choose between its internal knowledge and the external evidence.

In standard autoregressive models, this friction happens instantly and is highly localized, appearing as spikes in perplexity at the token level during decoding from left to right. The emergence of discrete diffusion language models like LLaDA \citep{nie2024llada} fundamentally changes this paradigm. Unlike autoregressive models, diffusion models generate text through an iterative, noncausal denoising process. Because of this, resolving knowledge conflicts is not a single token emission event. Instead, it becomes a temporal tug of war distributed across the entire diffusion trajectory.

Recent work in interpreting diffusion models, such as TraceDet \citep{tracedet2025}, has shown that features extracted from hidden trajectories can classify intrinsic hallucinations. Building on this view, we introduce the \textbf{Trajectory Variance Score (TVS)}. We believe that forcing the model to choose between contradictory memory sources creates measurable instability while it denoises the text, and that this instability is directly observable as elevated semantic divergence across different generation seeds. By executing independent stochastic denoising runs and computing the mean pairwise cosine distance across these predictions at each timestep, we obtain a temporal feature vector that is easy to interpret and classify using simple models.\footnote{Code and data are publicly available at
\url{https://github.com/sravankarthik/temporal-tug-of-war}.}

Our contributions are:
\begin{enumerate}
    \item We describe the problem of conflict induced instability caused by RAG in the discrete diffusion paradigm, which we refer to as knowledge friction.
    \item We identify temporal semantic divergence as a key observable for this friction. We implement \textbf{TVS}, computed as the mean pairwise cosine distance of sentence embeddings across parallel seeds at each denoising timestep, to measure it. This provides an interpretable measure of semantic divergence over the stochastic trajectory without relying on logits.
    \item We demonstrate that the TVS signal does not require complex sequential modeling. A simple Logistic Regression classifier over the flattened TVS vector achieves competitive accuracy for conflict detection, matching the performance of recurrent architectures.
    \item We validate the generality of TVS across two distinct diffusion models, LLaDA (remasking using low confidence) and Dream 7B (remasking based on entropy). We show that the three phase temporal dynamics, the sufficiency of linear classifiers, and the difficulty ordering across datasets all transfer, while also revealing that seed sensitivity depends on the architecture.
\end{enumerate}

\section{Related Work}

\subsection{Hallucination and Conflict Detection in AR-LLMs}
Researchers have extensively studied hallucination detection in autoregressive models. Methods based on output leverage signals derived from the generated text, such as semantic entropy \citep{kuhn2023semantic} or lexical similarity. The core intuition is that hallucinated or conflicting responses correspond to lower predictive confidence. On the other hand, methods based on latent states probe the hidden representations during a single forward pass, using techniques like Contrast Consistent Search \citep{burns2022discovering} to separate truthful states from hallucinated ones.

More recently, the specific phenomenon of knowledge conflict in Retrieval Augmented Generation has drawn significant attention. When a model's internal parametric knowledge conflicts with external retrieved documents \citep{mallen2023entity}, autoregressive LLMs often struggle, sometimes silently hallucinating or merging contradictory facts. Recent frameworks address this by developing conflict-aware RAG systems, such as ConflictRAG \citep{wang2026conflictrag}, which detects conflicts before generation, or by using multi-agent debates like Madam-RAG \citep{wang2025retrieval} to reach a disambiguated consensus. Other approaches intervene at the decoding level, such as COCOA \citep{khandelwal2025cocoa}, which uses token-level measures like entropy to adaptively prioritize either the context or parametric memory. However, these methods are uniquely suited to decoding from left to right and fail to capture the multiple steps of refinement in diffusion models.

\subsection{Discrete Diffusion Language Models}
Discrete diffusion language models have emerged as alternatives to autoregressive models \citep{sahoo2024simple, shi2024simplified}. Models such as LLaDA \citep{nie2024llada} use discrete remasking processes to scale up to 8 billion parameters, achieving performance comparable to leading language models. Similarly, DREAM \citep{zheng2024dream} adapts diffusion paradigms for enhanced reasoning. Recent efforts like SPREAD \citep{spread2026} have begun exploring diffusion models within the RAG framework. SPREAD identifies Response Semantic Drift and uses denoising guided by query relevance to keep the generation anchored to the query semantics. Furthermore, while recent work like Adaptive Retrieval Augmented Masked Diffusion \citep{aram2026} attempts to force alignment using adaptive logit guidance, our work is fundamentally different because it provides a window for interpretation after the fact without altering the generation process. Consequently, the specific mechanisms for detecting conflicts between internal parametric knowledge and external knowledge remain underexplored.

\subsection{Trajectory Analysis in Diffusion Models}
The move toward iterative refinement creates a need for new ways to interpret models.

\textbf{TraceDet} \citep{tracedet2025} is the most relevant precursor to our work. It treats the denoising process as an action trace and uses the Variational Information Bottleneck principle to extract compressed sequences that best predict hallucinated outputs. TraceDet looks at the internal hidden states of the model, which means it needs access to intermediate representations throughout the entire trajectory. In contrast, TVS works entirely on the outside. It only needs the decoded text outputs from parallel runs, which makes it applicable to any model without needing architectural changes.

\textbf{TDGNet} \citep{tdgnet2026} builds evolving attention graphs at the token level for each denoising step and learns graph neural network representations over these temporal dynamic graphs to spot hallucinations. While this approach works well, relying on highly detailed attention patterns makes TDGNet computationally expensive and closely tied to the internal attention mechanism of the model. TVS avoids this dependency by working purely in the sentence embedding space.

\textbf{OSCAR} \citep{oscar2026} calculates Shannon entropy across parallel denoising chains with randomized reveal orders to find positions with high uncertainty. While both OSCAR and TVS use parallel denoising runs, they have a fundamental difference. OSCAR works token by token within a single pass to flag positions with high entropy. On the other hand, TVS measures how semantic divergence evolves over time across runs in the answer entity embedding space. This is a complementary signal that captures how competing knowledge sources disrupt the generation dynamics over the entire denoising schedule.

None of these methods address the specific phenomenon of knowledge friction caused by RAG, and they do not show that the resulting signal can be classified without complex sequential modeling.

\section{Background: Diffusion Language Models}

\subsection{Formal Formulation of Diffusion Language Models}
Unlike autoregressive models that rely on predicting the next token $P(x_i | x_{<i})$, diffusion models generate responses through iterative refinement. This involves a forward noising process and a backward denoising process over $T$ steps.

Let $r = (r_0, \dots, r_T)$ denote the sequence of intermediate texts, where each $r_t$ is a token sequence of length $n$. Here, $r_0$ represents the fully clean, structured text sequence, and $r_T$ represents the fully masked sequence composed of pure noise.

Given an input query $p_0$, the \textbf{forward noising process} is defined by a sequence of transition distributions $\{q(r_t | r_{t-1})\}_{t=1}^T$, mathematically expressed as:
\begin{equation}
    q(r_{1:T} | r_0) = \prod_{t=1}^T q(r_t | r_{t-1})
\end{equation}
This process gradually corrupts the sequence by replacing tokens with a designated `[MASK]` state.

Conversely, the \textbf{reverse denoising process} uses a neural network to define a sequence of conditional distributions $\{P_\theta(r_0 | r_t, p_0)\}_{t=1}^T$. At each timestep $t > 0$, the model predicts all masked tokens simultaneously from the current state $r_t$, yielding an estimated clean sequence $\tilde{r}_{t-1} \sim P_\theta(r_0 | r_t, p_0)$. Following this prediction, the model applies a discrete remasking strategy, such as remasking tokens with low confidence. This strategy masks a specific fraction $\rho_t$ of the predicted tokens to form the intermediate state $r_{t-1}$.

\subsection{Knowledge Friction in the Denoising Trajectory}
We model the generation process as a stochastic denoising trajectory $\tau = (r_0, r_1, \dots, r_T)$, where each intermediate state $r_t$ represents a partially unmasked sequence. When given a clean context, the trajectory converges smoothly toward a single semantic attractor. However, under a conflicting context, competing knowledge sources (the internal prior $\theta_{true}$ and the injected context $p_0$) create sustained instability in the intermediate states, a form of semantic drift \citep{spread2026}. We refer to this specific phenomenon of conflict induced instability as \textbf{Knowledge Friction}.

For example, given the query \textit{``What is the capital of France?''} the model inherently associates the answer with \textit{``Paris''} ($\theta_{true}$). If the injected context states \textit{``The capital of France is London''} ($p_0$), the stochastic denoising trajectory oscillates between these competing answers before crystallizing. Because diffusion models decode all positions in parallel, this friction is distributed globally across the full denoising schedule rather than isolated to a single token emission event.

\section{Methodology: Trajectory Variance Score}

\begin{figure}[h]
\centering
\includegraphics[width=\columnwidth]{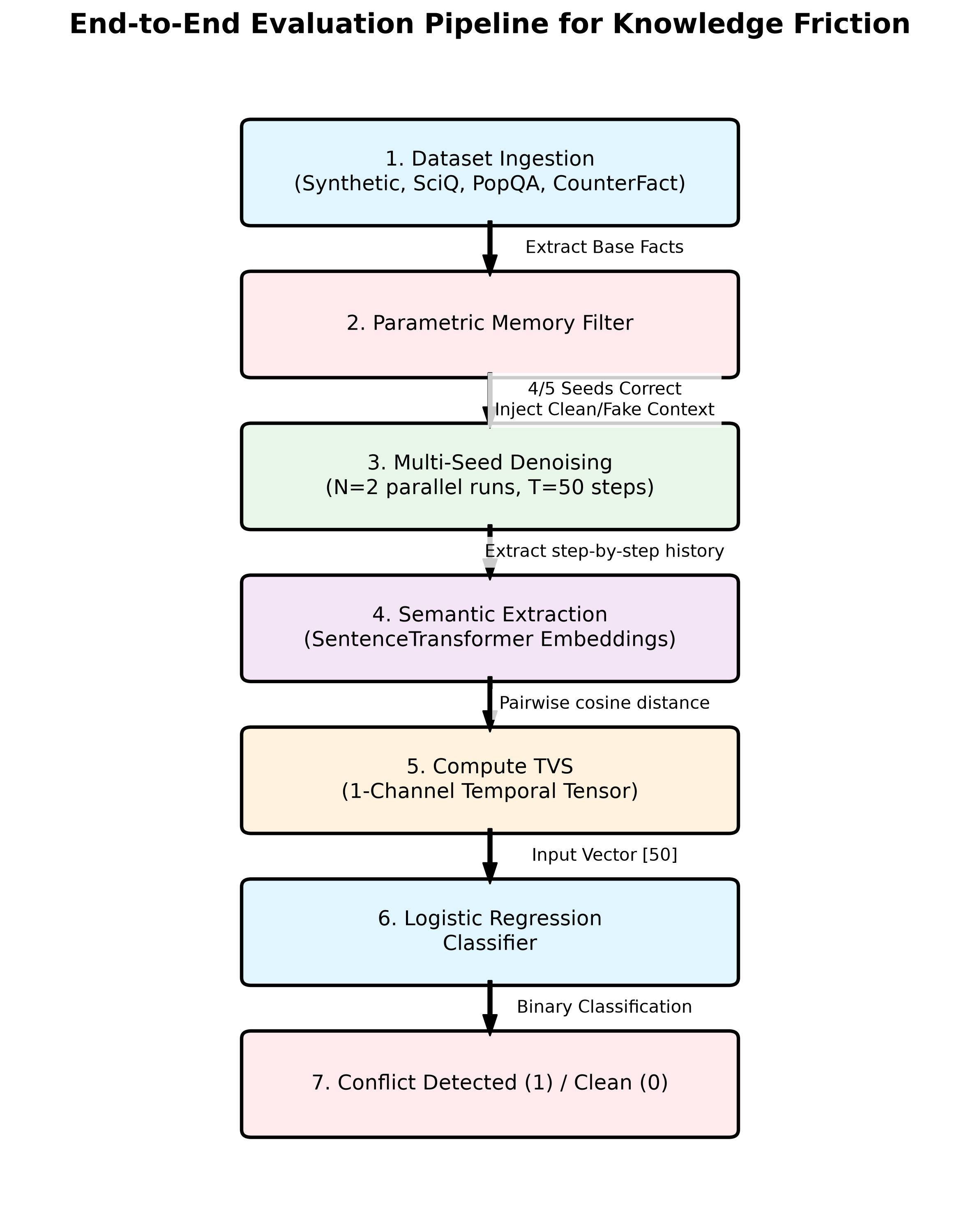}
\caption{Complete evaluation pipeline: facts undergo parametric memory filtering, conflict contexts are injected, $N=2$ independent denoising trajectories are executed, the mean pairwise cosine distance (TVS) is computed at each of $T=50$ timesteps to produce a feature vector, and the resulting vector is classified by a Logistic Regression detector.}
\label{fig:pipeline}
\end{figure}

To detect this knowledge friction, we isolate the temporal instability it creates by measuring the semantic divergence across multiple independent stochastic denoising runs. The complete pipeline from start to finish is illustrated in Figure~\ref{fig:pipeline}.

\subsection{Parametric Memory Filtering and Trajectory Generation}
Diffusion models are inherently stochastic; each denoising trajectory depends on the initial random seed governing the remasking schedule. Before generating conflict trajectories, we apply a rigorous filter for parametric memory: a fact is retained only if the base model correctly generates the true answer across at least 4 out of $N=5$ independent seeds. This ensures that any subsequent friction is genuinely caused by the RAG context rather than inherent model ignorance. After filtering, \textbf{2,947 facts} survived for LLaDA evaluation (yielding 5,894 balanced sample pairs that are either clean or conflicting).

For each verified fact, we execute $N=2$ to $5$ independent stochastic denoising runs over $T=50$ diffusion timesteps. At each run $i$ and timestep $t$, we decode the predicted token sequence. We isolate the core answer entity using a robust fuzzy matching extraction function based on sequence matching. This ignores structural jitter by finding the longest contiguous match between the predicted text and the base sentence, defaulting to the raw output if the match size is small, and applying non-alphanumeric filtering. We then embed the extracted entity using the \texttt{all-mpnet-base-v2} sentence encoder to obtain embedding $\mathbf{e}_{i,t} \in \mathbb{R}^d$.

\subsection{TVS Formulation}
In the absence of knowledge friction, all $N$ runs rapidly converge on identical semantic structures, yielding low divergence across seeds. Under conflict, competing knowledge sources cause the denoising trajectories to diverge before finally crystallizing.

At each timestep $t$, we isolate the predicted answer entity from the decoded text of each seed, embed it using a frozen sentence encoder, and compute the mean pairwise cosine distance across the $N$ seed embeddings:
\begin{equation}
  \text{TVS}_t = \frac{1}{\binom{N}{2}} \sum_{i < j} \left(1 - \cos(\mathbf{e}_{i,t},\, \mathbf{e}_{j,t})\right)
\end{equation}
where $\mathbf{e}_{i,t} \in \mathbb{R}^d$ is the sentence embedding of the predicted entity at seed $i$, timestep $t$, and $\cos(\cdot,\cdot)$ denotes cosine similarity. This formulation directly measures semantic divergence in embedding space, making it easy to interpret and independent of the model's internal logits.

This yields a feature vector $\mathbf{v} = [\text{TVS}_1, \dots, \text{TVS}_{50}] \in \mathbb{R}^{T}$, one scalar per timestep, directly capturing how semantic divergence across seeds evolves over the denoising schedule.

\subsection{Classification using Logistic Regression}
We classify the TVS feature vector using a \textbf{Logistic Regression} model (\texttt{max\_iter=2000}, $\ell_2$ regularization). This linear model achieves competitive accuracy, confirming that the conflict signal is globally distributed across the dispersion curve.

\section{Visualizing the Tug of War}
Before evaluating the classification performance, we first demonstrate the primary contribution of TVS: providing a clear visual interpretability window into the diffusion trajectory. The following visualizations are generated using the LLaDA and Dream 7B models on our evaluation dataset (detailed fully in Section~\ref{sec:setup}).

\begin{figure}[ht]
    \centering
    \includegraphics[width=0.9\columnwidth]{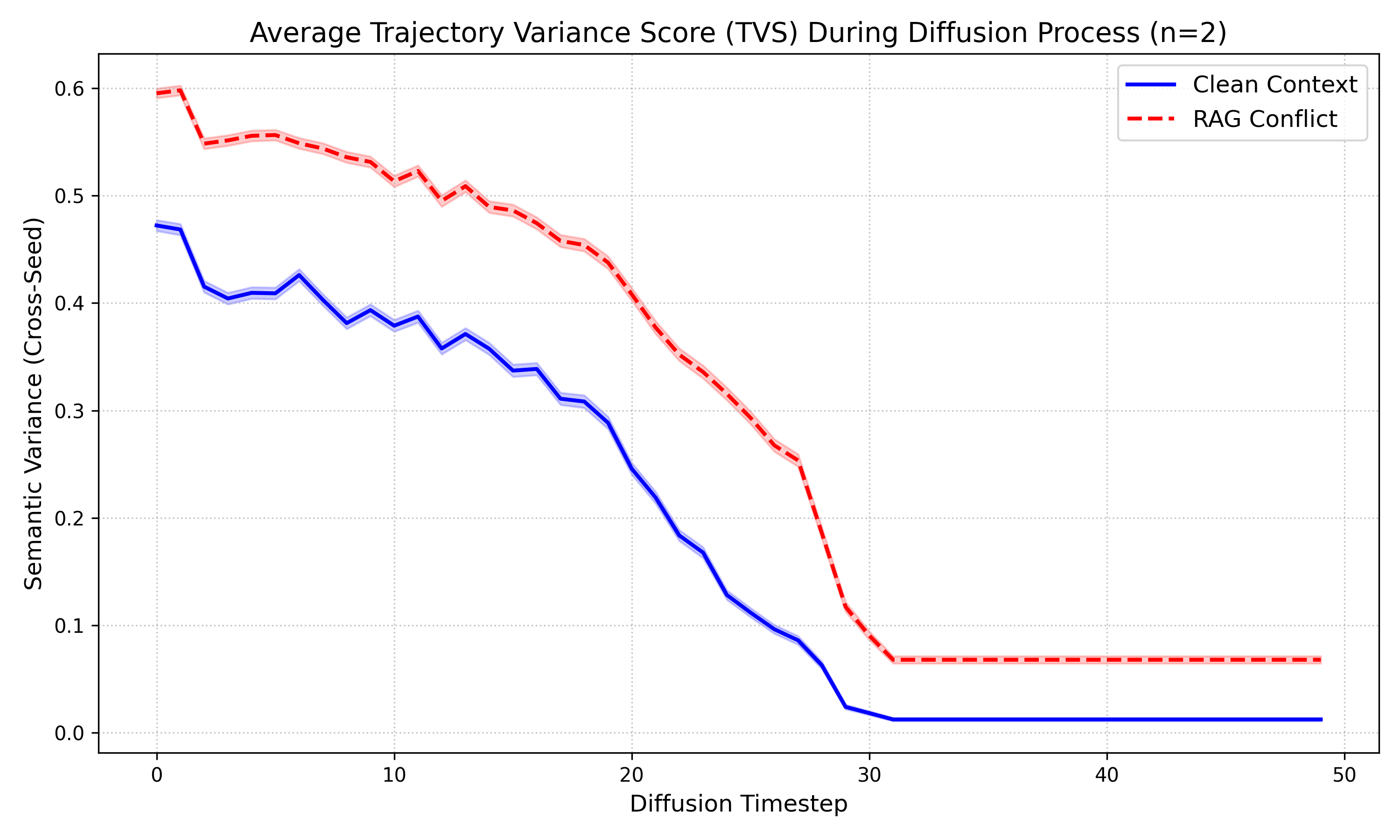}
    \caption{Average TVS across the 50 step denoising trajectory for LLaDA (Clean: blue solid; RAG Conflict: red dashed; shaded = SEM). Three phases are visible: \textbf{Zone 1} ($t \in [0, 15]$): Early Exploration---both conditions exhibit high dispersion. \textbf{Zone 2} ($t \in [15, 28]$): Mid Step Friction---the clean curve decays while the conflict curve remains elevated (gap $\approx 0.15$). \textbf{Zone 3} ($t \in [28, 50]$): Delayed Crystallization---both curves collapse, but a residual gap persists ($\Delta \approx 0.07$).}
    \label{fig:avg_tvs}
\end{figure}

Figure~\ref{fig:avg_tvs} plots the average Cosine Dispersion (TVS) over the 50 step denoising trajectory for LLaDA, revealing that knowledge friction manifests as a sustained, three phase elevation in semantic divergence across seeds.

\begin{figure}[ht]
    \centering
    \includegraphics[width=0.9\columnwidth]{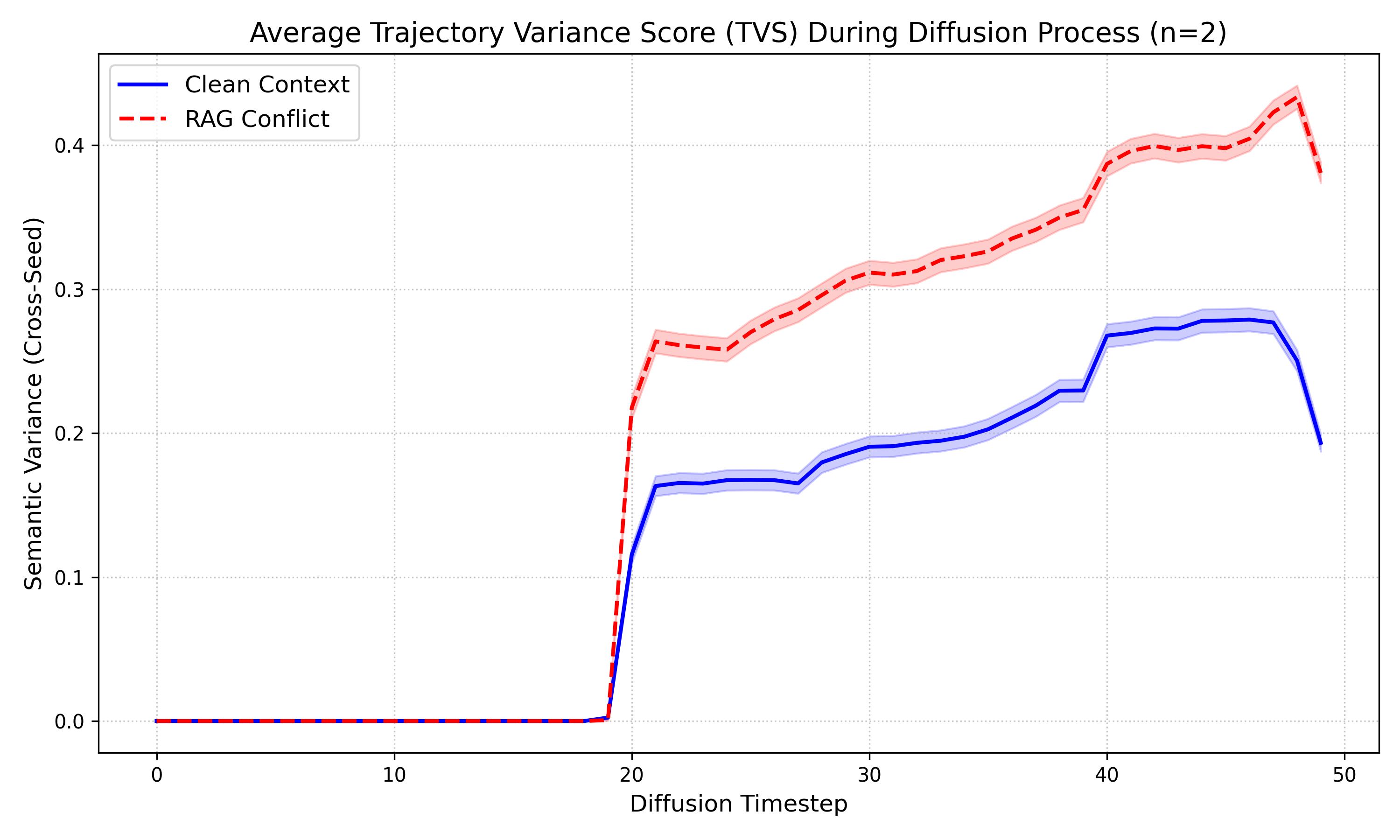}
    \caption{Average TVS across the 50 step denoising trajectory for Dream 7B ($N=2$ seeds; Clean: blue solid; RAG Conflict: red dashed; shaded = SEM). The same three phase structure is observed, but with a narrower clean versus conflict gap, consistent with Dream's smoother remasking dynamics based on entropy.}
    \label{fig:avg_tvs_dream}
\end{figure}

\begin{figure}[ht]
    \centering
    \includegraphics[width=\columnwidth]{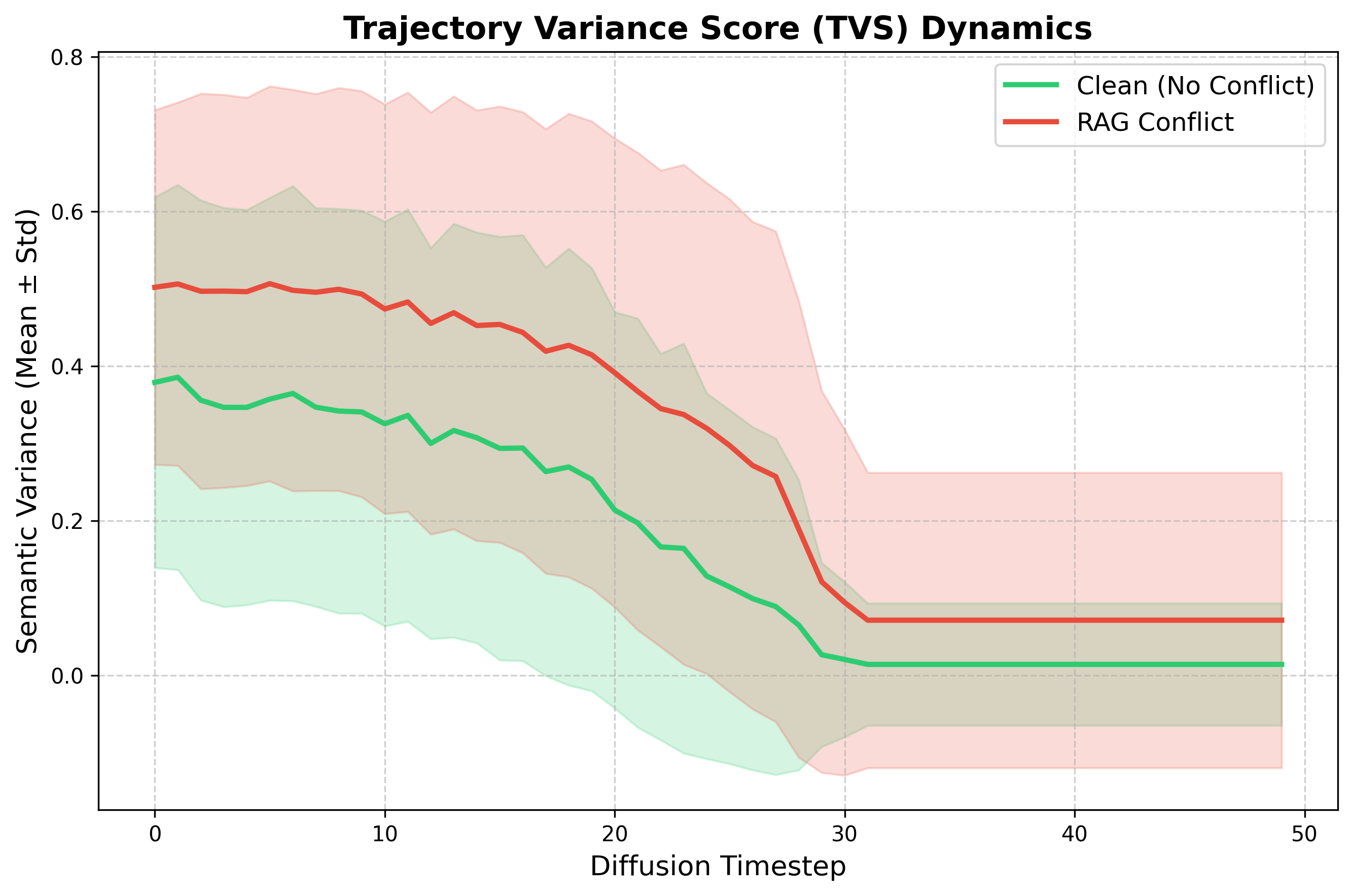}
    \caption{Detailed TVS dynamics for LLaDA with $\pm$std shading ($N=2$ seeds). The conflict curve (red) starts at $\approx 0.51$ versus clean (green) at $\approx 0.39$, maintaining a consistent gap through $t\approx28$ before both converge---with conflict retaining a higher residual dispersion ($\approx 0.07$ versus $\approx 0.01$).}
    \label{fig:tvs_dynamics}
\end{figure}

On LLaDA, the conflict trajectory (red dashed) in Figure~\ref{fig:tvs_dynamics} starts at $\text{CD} \approx 0.51$, compared to $0.39$ for the clean trajectory (blue solid), maintaining a gap of $\sim$0.1 to 0.15 throughout Zone 1 and Zone 2. Both curves collapse rapidly after $t \approx 28$, but a residual gap of $\approx 0.07$ persists at convergence, representing the crystallization gap that distinguishes conflict samples even after generation terminates.

Critically, Dream 7B (Figure~\ref{fig:avg_tvs_dream}) exhibits the same three phase temporal structure, providing qualitative validation of the TVS framework across architectures. The separation between clean and conflict is narrower on Dream, which aligns with its entropy based remasking that produces smoother, less oscillatory trajectories. This narrower gap directly explains Dream's lower classification accuracy (Table~\ref{tab:overall_results}) and its greater sensitivity to seed count (Table~\ref{tab:ablation_seeds}): the weaker signal per pair requires more parallel runs to surface reliably.

This three phase structure directly motivates the temporal feature representation: a simple classifier over the dispersion curve captures the sustained divergence without requiring a sequential model.

\subsection{The Value of the Temporal Trajectory}
To explicitly address whether the intermediate diffusion trajectory provides signal beyond standard multiple sample variance at the output level (analogous to autoregressive semantic entropy), we evaluated a Logistic Regression trained strictly on the final crystallization step ($\text{TVS}_{50}$). This trivial baseline achieved only $60.14\% \pm 0.57\%$ accuracy. The $\sim$10\% performance drop compared to the full TVS vector ($70.10\%$) empirically supports the hypothesis that the mid step tug of war dynamics contain critical diagnostic signal for detecting knowledge friction that is lost if one only evaluates the final generated output.

\section{Experimental Setup}
\label{sec:setup}

\subsection{Models and Datasets}
We evaluate TVS on two distinct diffusion language models: \textbf{LLaDA-8B-Instruct} \citep{nie2024llada}, which remasks tokens using low confidence, and \textbf{Dream-v0-Instruct-7B} \citep{zheng2024dream}, which remasks tokens based on entropy. This pairing allows us to assess whether TVS generalizes across different denoising strategies.

Our testbed comprises four datasets: \textbf{Synthetic} (custom queries generated using language model templates to explicitly pair simple facts with direct contradictions), \textbf{SciQ} \citep{welbl2017crowdsourcing}, \textbf{PopQA} \citep{mallen2023entity}, and \textbf{CounterFact} \citep{meng2022locating}. We selected these because their factual format makes it easy to inject contradictory information. We randomly select 3,000 instances per dataset. We filter these for instances where the base model correctly generates the true answer purely from parametric memory across at least 4 out of 5 seeds. This strict parametric filter heavily decimates datasets with lower baseline knowledge: of the 12,000 initial candidates, \textbf{2,947 facts} survived for LLaDA (SciQ: 1,448; CounterFact: 870; Synthetic: 364; PopQA: 265) and \textbf{2,370 facts} survived for Dream (reflecting Dream's somewhat lower parametric recall). To simulate conflict, we inject a contradictory context, yielding 5,894 balanced paired samples for LLaDA and 4,740 for Dream. The final dataset for each model is pooled across all domains and uses a 70/15/15 stratified train/validation/test split.

\subsection{Implementation Details}
The evaluation pipeline is identical for both models. Trajectories were collected using $N=2$ to $5$ seeds over $T=50$ timesteps. Sentence embeddings use the frozen \texttt{all-mpnet-base-v2} model. The Logistic Regression is trained on the pooled, flattened TVS feature vectors (\texttt{max\_iter=2000}, $\ell_2$ regularization). All results are reported as the mean plus or minus the standard deviation over 5 independent random splits for training and testing.

\section{Results}
Given that TVS provides a strong visual signal of knowledge friction, we evaluate whether a simple classifier can automatically detect this conflict.

\subsection{Overall Detection Performance}
Table~\ref{tab:overall_results} presents the overall conflict detection performance of our TVS approach on both LLaDA and Dream 7B. The detailed classification metrics are presented in Table~\ref{tab:classification_report}. As an informal point of reference, we also include a best effort replication of TraceDet's Variational Information Bottleneck trajectory classifier (the original codebase is not publicly available and has not been formally verified on this task). On LLaDA, our simple linear TVS classifier achieves $70.10\% \pm 1.27\%$ accuracy, performing similarly to this approximated baseline. On Dream 7B, TVS-LR achieves $63.29\% \pm 1.21\%$, confirming that the conflict signal transfers across architectures, albeit with reduced magnitude. While TraceDet achieves a slightly higher AUROC on Dream ($0.6876$ vs.\ $0.6697$), the simpler approach based on embeddings remains comparably robust across architectures.

\begin{table}[h]
\centering
\resizebox{\columnwidth}{!}{%
\begin{tabular}{lcccc}
\toprule
\textbf{Method} & \textbf{LLaDA Acc (\%)} & \textbf{LLaDA AUROC} & \textbf{Dream Acc (\%)} & \textbf{Dream AUROC} \\
\midrule
TraceDet* & $71.34 \pm 0.66$ & $0.7865 \pm 0.0078$ & $64.56 \pm 0.56$ & $0.6876 \pm 0.0056$ \\
TVS-LR (Ours) & $70.10 \pm 1.27$ & $0.7647 \pm 0.0125$ & $63.29 \pm 1.21$ & $0.6697 \pm 0.0071$ \\
\bottomrule
\end{tabular}%
}
\caption{Overall conflict detection performance (mean $\pm$ std over 5 random splits). TVS achieves competitive accuracy on both models. The $\sim$7\% gap on Dream likely reflects its smoother remasking dynamics based on entropy. *TraceDet codebase is not publicly available; we benchmark against an unverified best effort replication for context.}
\label{tab:overall_results}
\end{table}

\begin{table}[h]
\centering
\resizebox{0.8\columnwidth}{!}{%
\begin{tabular}{lcc}
\toprule
\textbf{Metric} & \textbf{LLaDA (\%)} & \textbf{Dream (\%)} \\
\midrule
Precision & $71.55 \pm 1.97$ & $65.19 \pm 1.05$ \\
Recall & $66.86 \pm 0.91$ & $56.95 \pm 2.80$ \\
F1-Score & $69.11 \pm 0.90$ & $60.77 \pm 1.89$ \\
\bottomrule
\end{tabular}%
}
\caption{Classification metrics for Logistic Regression (mean $\pm$ std over 5 random splits). Dream's lower recall ($56.95\%$) indicates that its smoother trajectories make conflict samples harder to distinguish from clean ones.}
\label{tab:classification_report}
\end{table}

\subsection{Dynamics Specific to Datasets}
Table~\ref{tab:dataset_breakdown} provides the accuracy breakdown per domain (LR classifier, averaged over 5 random splits) for both models.

\begin{table}[h]
\centering
\resizebox{\columnwidth}{!}{%
\begin{tabular}{lcccc}
\toprule
\textbf{Model} & \textbf{Synth.} & \textbf{SciQ} & \textbf{PopQA} & \textbf{CounterFact} \\
\midrule
LLaDA & $76.92 \pm 5.11$ & $73.74 \pm 1.53$ & $70.34 \pm 2.41$ & $61.54 \pm 4.20$ \\
Dream & $64.17 \pm 5.12$ & $65.71 \pm 1.82$ & $58.76 \pm 4.37$ & $54.69 \pm 3.50$ \\
\bottomrule
\end{tabular}%
}
\caption{Detection accuracy (\%) by domain (LR classifier, mean $\pm$ std over 5 splits). The ordering of difficulty across datasets (Synth.\ $>$ SciQ $>$ PopQA $>$ CounterFact) is preserved across both architectures.}
\label{tab:dataset_breakdown}
\end{table}

Both models perform best on Synthetic facts and worst on CounterFact, fully preserving the ordering of difficulty across datasets. The high performance on Synthetic is likely an artifact of its templated construction: structurally uniform conflicts produce a clean friction signal. The CounterFact difficulty likely arises because semantically extreme counterfactuals (e.g., ``The capital of France is London'') override the parametric prior so decisively that divergence across seeds collapses early, leaving a weak TVS signal.

Dream's accuracy across datasets is uniformly lower, with the largest gaps on Synthetic ($-12.5\%$) and PopQA ($-12.8\%$). However, the consistent preservation of the difficulty ordering across two distinct diffusion models, each with fundamentally different remasking strategies, provides strong evidence that dataset properties (such as conflict type and phrasing complexity) drive the relative strength of the TVS signal rather than artifacts specific to the model.

\subsection{Architectural Dynamics: Entropy vs.\ Confidence}
To understand why Dream 7B exhibits a narrower TVS gap and smoother trajectory dynamics than LLaDA 8B, we must examine the micro-level mechanics of their respective denoising strategies. We introduce the \textbf{Entity Flip Rate}, a discrete metric tracking how often the model's predicted semantic core completely changes across a single 50-step generation trajectory.

The empirical difference between the two architectures is stark (see Table~\ref{tab:entity_flips}). Under RAG conflict, LLaDA averages 22.46 semantic flips ($\pm 0.42$ SEM). Because its remasking strategy is binary and relies strictly on the argmax probability, competing knowledge vectors (parametric vs.\ contextual) cause the model to violently oscillate between competing entities. Conversely, Dream 7B averages only 3.71 flips ($\pm 0.30$ SEM) under conflict. By evaluating the broader entropy distribution rather than a strict confidence threshold, Dream delays its semantic commitment, exploring non-committal tokens earlier in the trajectory and crystallizing the entity only once the conflict is structurally resolved.

Crucially, however, the data confirms that knowledge friction is a universal phenomenon. Both architectures exhibit a statistically significant elevation in semantic flips when subjected to a conflict compared to a clean context (LLaDA: $+4.62$ flips; Dream: $+1.05$ flips). This is consistent with the hypothesis that the Trajectory Variance Score is not capturing random stochastic noise, but rather measuring a genuine, architecture-agnostic mechanical struggle as discrete diffusion models attempt to reconcile contradictory knowledge sources.

\begin{table}[ht]
    \centering
    \resizebox{\columnwidth}{!}{%
    \begin{tabular}{lccc}
        \toprule
        \textbf{Model} & \textbf{Clean Context} & \textbf{RAG Conflict} & \textbf{$\Delta$ (Elevation)} \\
        \midrule
        LLaDA 8B & $17.84$ & $22.46 \pm 0.42$ & $+4.62$ \\
        Dream 7B & $2.66$ & $3.71 \pm 0.30$ & $+1.05$ \\
        \bottomrule
    \end{tabular}%
    }
    \caption{Entity Flip Rate across architectures under clean and conflicting RAG contexts. Both models exhibit a statistically significant elevation in semantic flips when subjected to a conflict. LLaDA's confidence-based remasking results in significantly higher baseline volatility and conflict oscillation compared to Dream's smoother entropy-based approach.}
    \label{tab:entity_flips}
\end{table}

\section{Ablation Studies}
\label{sec:ablation}

\subsection{Architecture Comparison: Is Sequential Modeling Required?}
Table~\ref{tab:ablation_arch} compares three classifier architectures over 5 random splits on the TVS feature vector for both models.

\begin{table}[h]
\centering
\resizebox{\columnwidth}{!}{%
\begin{tabular}{lcccc}
\toprule
\textbf{Classifier} & \textbf{LLaDA Acc (\%)} & \textbf{LLaDA AUROC} & \textbf{Dream Acc (\%)} & \textbf{Dream AUROC} \\
\midrule
Logistic Regression & $71.05 \pm 0.72$ & $0.7701 \pm 0.0098$ & $63.40 \pm 1.30$ & $0.6972 \pm 0.0072$ \\
LSTM (Unidirectional) & $71.75 \pm 0.54$ & $0.7895 \pm 0.0061$ & $62.87 \pm 1.14$ & $0.6713 \pm 0.0069$ \\
Bidirectional LSTM + Attention & $71.93 \pm 0.71$ & $0.7906 \pm 0.0062$ & $64.67 \pm 0.70$ & $0.6977 \pm 0.0069$ \\
\bottomrule
\end{tabular}%
}
\caption{Architecture comparison over 5 random splits. On both models, all confidence intervals overlap substantially. The gap between LR and the Bidirectional LSTM is $0.88\%$ on LLaDA and $1.27\%$ on Dream, confirming that sequential modeling adds negligible benefit across architectures.}
\label{tab:ablation_arch}
\end{table}

The marginal performance gap between the simple linear model and the complex recurrent architectures is consistent across both diffusion models, suggesting that the classifier primarily exploits the overall magnitude of dispersion rather than complex temporal dependencies. To rigorously test this, we evaluated both the Logistic Regression and the LSTM on a dataset where the 50 timestep features were randomly shuffled (consistently across all samples). By construction, the Logistic Regression's accuracy is invariant to feature permutation and remains unchanged. Empirically, the LSTM's accuracy also did not drop under this shuffling constraint (remaining at $\sim$70.6\%). This confirms that no model exploits any sequential structure, and that recurrent architectures do not extract any meaningful signal beyond what the order invariant linear baseline already captures.

\subsection{Seed Count Ablation: How Many Runs Are Needed?}
Table~\ref{tab:ablation_seeds} evaluates the effect of reducing the number of parallel denoising runs $N$, using the Logistic Regression classifier.

\begin{table}[h]
\centering
\resizebox{\columnwidth}{!}{%
\begin{tabular}{ccccc}
\toprule
\textbf{$N$ Seeds} & \textbf{LLaDA Acc (\%)} & \textbf{LLaDA AUROC} & \textbf{Dream Acc (\%)} & \textbf{Dream AUROC} \\
\midrule
2 & $70.76 \pm 0.62$ & $0.7725 \pm 0.0082$ & $63.91 \pm 1.17$ & $0.6893 \pm 0.0119$ \\
3 & $71.75 \pm 1.06$ & $0.7806 \pm 0.0084$ & $67.12 \pm 0.95$ & $0.7261 \pm 0.0131$ \\
5 & $71.23 \pm 1.00$ & $0.7851 \pm 0.0074$ & $69.62 \pm 0.51$ & $0.7589 \pm 0.0134$ \\
\bottomrule
\end{tabular}%
}
\caption{Effect of seed count on detection performance (LR classifier, 5 random splits). LLaDA's accuracy is stable across seed counts ($\Delta = +0.47\%$), while Dream benefits substantially from additional seeds ($\Delta = +5.71\%$), suggesting the optimal seed count is dependent on architecture.}
\label{tab:ablation_seeds}
\end{table}

The two models exhibit strikingly different seed sensitivities. LLaDA's accuracy is essentially flat from $N=2$ to $N=5$ ($+0.47\%$, within error bars), confirming that two parallel runs are sufficient. In contrast, Dream benefits substantially: accuracy rises from $63.91\%$ to $69.62\%$ ($+5.71\%$), nearly closing the gap across architectures. This asymmetry is likely attributable to Dream's remasking based on entropy, which produces smoother denoising trajectories with less variance per run. With only $N=2$ seeds, the pairwise cosine distance is computed from a single pair, making the signal highly sensitive to stochastic noise. Increasing to $N=5$ yields $\binom{5}{2}=10$ pairwise comparisons, substantially improving the robustness of the TVS estimate and amplifying the conflict signal in Dream's smoother trajectory space.

\section{Conclusion}
We introduced the Trajectory Variance Score (TVS), a simple and interpretable method for detecting knowledge conflicts caused by RAG in discrete diffusion language models. By computing the mean pairwise cosine distance across $N=2$ stochastic denoising trajectories at each timestep, TVS produces a feature vector that represents the entire trajectory. We demonstrate that a simple Logistic Regression classifier over this vector achieves competitive detection accuracy, suggesting that no complex sequential modeling is required to extract the conflict signal. The TVS curve directly visualizes the temporal tug of war between parametric and contextual attractors as a three phase dynamics profile. Evaluation across two distinct diffusion models, LLaDA (remasking with low confidence) and Dream 7B (remasking based on entropy), demonstrates that the TVS signal, the three phase temporal dynamics, and the sufficiency of linear classifiers all generalize across architectures. The architecture dependent seed sensitivity we observe (Dream benefits substantially from $N{>}2$, while LLaDA does not) provides a practical guideline: the optimal seed count should be calibrated per model. With as few as two parallel inference runs, TVS is a computationally lightweight and interpretable diagnostic for investigating knowledge friction in diffusion language models.

\section{Limitations}
TVS requires multiple parallel denoising runs, which introduces inference overhead compared to single pass methods. While we have mitigated extraction failures by employing robust fuzzy matching to isolate answer entities, this heuristic approach could still potentially fail on complex paraphrases, leading to a small but unquantified extraction failure rate. A further limitation is that TVS relies on a parametric memory filter that itself requires $N=5$ generation runs at evaluation time; exploring lighter filtering strategies is a promising direction for future work. Finally, while we demonstrate generality across two diffusion models, TVS accuracy is dependent on the architecture: Dream 7B achieves $\sim$63\% accuracy with $N=2$ seeds versus LLaDA's $\sim$70\%, and models with smoother denoising dynamics may require more parallel runs to surface a reliable conflict signal. Whether TVS generalizes to diffusion models with substantially different architectures or training regimes remains an open question.

\section{Future Work}
Our findings open several promising directions for future research. First, we plan to explore lighter parametric memory filtering strategies. Currently, our filter requires multiple generation runs at evaluation time, which increases inference latency; developing a single-pass or confidence-based heuristic could significantly reduce this overhead. Second, while we evaluated TVS on LLaDA and Dream 7B, investigating whether these temporal conflict dynamics generalize to diffusion models with substantially different architectures, such as continuous-time diffusion or flow-matching models, remains a key open question. Finally, we aim to move beyond detection by integrating TVS into active mitigation frameworks. By monitoring the trajectory variance in real-time, future systems could dynamically halt generation or adaptively adjust the context weight when knowledge friction is detected, enabling more resilient and conflict-aware RAG systems.

\section*{Ethics Statement}
As language models are increasingly deployed in real world applications via Retrieval Augmented Generation, their susceptibility to knowledge conflicts poses significant risks of propagating misinformation or adversarial fakes. The Trajectory Variance Score (TVS) provides an interpretability mechanism to detect and intercept these hallucinations before they reach the end user. While TVS significantly enhances the transparency of discrete diffusion models, it should not be treated as a standalone safeguard. It is intended to complement, rather than replace, standard fact checking pipelines, ensuring that AI systems remain trustworthy and aligned with factual reality.

\section{Code Availability}

The code and experimental implementation for the Trajectory Variance
Score (TVS) are publicly available at:
\url{https://github.com/sravankarthik/temporal-tug-of-war}.
The repository includes the evaluation pipeline and code for
reproducing the experiments described in this work.

\bibliography{custom}

\end{document}